%% file: main.tex
\documentclass[runningheads]{llncs}
\usepackage{graphicx}
\usepackage{amsmath}
\usepackage{varwidth}

\usepackage{graphicx}

\usepackage{subfigure}
\usepackage{amsmath,amsfonts,bm}
\usepackage{amsfonts}
\usepackage{amssymb}
\usepackage{amsxtra}

\newcommand{\comment}[1]{}

\usepackage{multicol}
\usepackage{tabu}

\usepackage{adjustbox}

\usepackage{multirow}
\usepackage{flushend}

\usepackage{float}

\usepackage{caption}

\usepackage{calc}

\usepackage{cancel}

\usepackage{mathrsfs}
\usepackage{algorithm}
\usepackage{algorithmic}

\usepackage{cuted}
\usepackage{blindtext, graphicx}
\usepackage{amsmath,float}
\usepackage{dirtytalk}
\usepackage{url}
\usepackage{amsmath,amsfonts,bm}
\usepackage[utf8]{inputenc}
\usepackage[english]{babel}
\usepackage{suffix}
\usepackage{mathtools}
\usepackage{multicol}
\usepackage{tabu}

\usepackage{caption}

\usepackage{graphics}
\usepackage{adjustbox}
\usepackage{lipsum}
\usepackage{stfloats}
\usepackage{multirow}

\usepackage{flushend}
\usepackage{tikz,colortbl}
\usepackage{zref-savepos}
\usepackage{pifont}

\usetikzlibrary{calc}

\allowdisplaybreaks

\usepackage{soul}

\usepackage{color}

\begin{document}
\title{NNLander-VeriF: A Neural Network Formal Verification Framework for Vision-Based Autonomous Aircraft Landing\thanks{This work was supported by the National Science Foundation under grant numbers \#2002405 and \#2013824.}}
\titlerunning{NN Formal Verification Framework for Vision-Based Aircraft Landing}
%
\author{Ulices Santa Cruz\inst{1} \and
Yasser Shoukry\inst{1}}
\authorrunning{U. Santa Cruz and Y. Shoukry}
%
\institute{University of California Irvine, Irvine CA, USA 
\\
\email{usantacr@uci.edu},
\email{yshoukry@uci.edu}
}
\maketitle              
\begin{abstract}
In this paper, we consider the problem of formally verifying a Neural Network (NN) based autonomous landing system. In such a system, a NN controller processes images from a camera to guide the aircraft while approaching the runway. 
%
A central challenge for the safety and liveness verification of vision-based closed-loop systems is the lack of mathematical models that captures the relation between the system states (e.g., position of the aircraft) and the images processed by the vision-based NN controller. Another challenge is the limited abilities of state-of-the-art NN model checkers. Such model checkers can reason only about simple input-output robustness properties of neural networks. This limitation creates a gap between the NN model checker abilities and the need to verify a closed-loop system while considering the aircraft dynamics, the perception components, and the NN controller. To this end, this paper presents NNLander-VeriF, a framework to verify vision-based NN controllers used for autonomous landing. NNLander-VeriF addresses the challenges above by exploiting geometric models of perspective cameras to obtain a mathematical model that captures the relation between the aircraft states and the inputs to the NN controller. 
By converting this model into a NN (with manually assigned weights) and composing it with the NN controller, one can capture the relation between aircraft states and control actions using one augmented NN. Such an augmented NN model leads to a natural encoding of the closed-loop verification into several NN robustness queries, which state-of-the-art NN model checkers can handle.
Finally, we evaluate our framework to formally verify the properties of a trained NN and we show its efficiency.



\end{abstract}
%
%
%





\input{SEC1-Introduction/intro}
\input{SEC2-Problem/problem}
\input{SEC4-Framework/framework}
\input{SEC6-Numerical/example}
\input{SEC7-Conclusion/conclusion}

%
%
%
\bibliographystyle{splncs04}
\bibliography{mybibliography}
%




\end{document}

%% file: SEC1-Introduction/intro.tex
\section{INTRODUCTION}
Machine learning models, like deep neural networks, are used heavily to process high-dimensional imaging data like LiDAR scanners and cameras. These data driven models are then used to provide estimates for the surrounding environment which is then used to close the loop and control the rest of the system. Nevertheless, the use of such data-driven models in safety-critical systems raises several safety and reliability concerns. It is unsurprising the increasing attention given to the problem of formally verifying Neural Network (NN)-based systems.

The work in the literature of verifying NNs and NN-based systems can be classified into \emph{component-level} and \emph{system-level} verification. Representatives of the first class, namely \emph{component-level} verification are the work on creating specialized decision procedures that can reason about input-output properties of NNs~\cite{katz2019marabou,KatzReluplexEfficientSMT2017a,ehlers2017formal,BakImprovedGeometricPath2020,KhedrPEREGRiNNPenalizedRelaxationGreedy2020,ferlez2021fast,ferlez2020bounding,TranNNVNeuralNetwork2020,WangNeuralNetworkControl2020}. In all these works, the focus is to ensure that inputs of the NN that belong to a particular convex set will result in NN outputs that belong to a defined set of outputs. Such input-output specification allows designers to verify interesting properties of NN like robustness to adversarial inputs and verify the safety of collision avoidance protocols.
For a comparison between the details and performance of these NN model checkers, the reader is referred to the annual competition on verification of neural networks~\cite{vnn2020}. Regardless of the improvements observed every year in the literature of NN model checkers, verifying properties of perception and vision-based systems as a simple input-output property of NNs is still an open challenge.

On the other hand, \emph{system-level} verification refers to the ability of reasoning about the temporal evolution of the whole system (including the NNs) while providing safety and liveness assurance~\cite{cruz2021safe,sun2021provably,sun2019formal,fremont2020formal}. A central challenge to verify systems that rely on vision-based systems and other high-bandwidth signals (e.g., LiDARs) is the need to explicitly model the imaging process, i.e., the relation between the system state and the images created by cameras and LiDARs~\cite{sun2019formal}. While first steps were taken to provide formal models for LiDAR based systems~\cite{sun2019formal}, very little attention is given to perception and vision-based systems. In particular, current state-of-the-art aims to avoid modeling the perception system formally, and instead focus on the use of abstractions of the perception system~\cite{katz2021verification,hsieh2021verifying}. Unfortunately, these abstractions are only tested on a set of samples and lack any formal guarantees in their ability to model the perception system formally. Other techniques use the formal specifications to guide the generation of test scenarios to increase the chances of finding a counterexample but without the ability to formally prove the correctness of the vision-based system~\cite{fremont2020formal}.

Motivated by the lack of formal guarantees of the abstractions of perception components~\cite{katz2021verification,hsieh2021verifying}, we argue in this paper for the need to formally model such perception components. Fortunately, such models were historically investigated in the literature of machine vision before the explosion of using data-driven approaches in machine learning~\cite{ma2012invitation,faugeras1993three}. While these physical/geometrical models of perception were shown to be complex to design vision-based systems with high performance, we argue that these models can be used for verification. In other words, we employ the philosophy of data-driven design of vision-based systems and model-based verification of such systems.



In this paper, we employ our philosophy above to the problem of designing a vision-based NN that controls aircraft while approaching runways to perform autonomous landing. Such a problem enjoys geometric nature that can be exploited to develop a geometrical/physical model of the perception system, yet represent an important real-world problem of interest to the autonomous systems designers. In particular, we present NNLander-VeriF, a framework for formal verification of vision-based autonomous aircraft landing. This framework provides several contributions to the state of the art:
\begin{itemize}
    \item The proposed framework exploits the geometry of the autonomous landing problem to construct a formal model for the image formation process (a map between the aircraft states and the image produced by the camera). This formal model is designed such that it can be encoded as a neural network (with manually chosen weights) that we refer to as the perception NN. By augmenting the perception NN along with the NN controller (which maps camera images into control actions), we obtain a formal relation between the aircraft states and the control action that is amenable to verification.
    \item The proposed framework uses symbolic abstraction of the physical dynamics of the aircraft to divide the problem of model checking the system-level safety and liveness properties into a set of NN robustness queries (applied to the augmented NN obtained above). Such robustness queries can be carried out efficiently using state-of-the-art component-level NN model checkers.
    \item We evaluated the proposed framework on a NN controller trained using imitation learning. 
\end{itemize}

%% file: SEC2-Problem/problem.tex
\section{PROBLEM FORMULATION}

\textbf{Notation.}
We will denote by $\mathbb{N}$, $\mathbb{B}$, $\mathbb{R}$ and $\mathbb{R^+}$ the set of natural, Boolean, real, and non-negative real numbers, respectively. 
We use $||x||_{\infty}$ to denote the infinity norm of a vector $x\in\mathbb{R}^n$. Finally, we denote by $\mathcal{B}_r(c)$ the infinity norm ball centered at $c$ with radius $r$, i.e., $\mathcal{B}_r(c) = \{x \in \mathbb{R}^n \; | \;\; || c - x||_{\infty} \le r\}$.

\begin{figure}[t!]
\centering
\includegraphics[width=1.0\columnwidth]{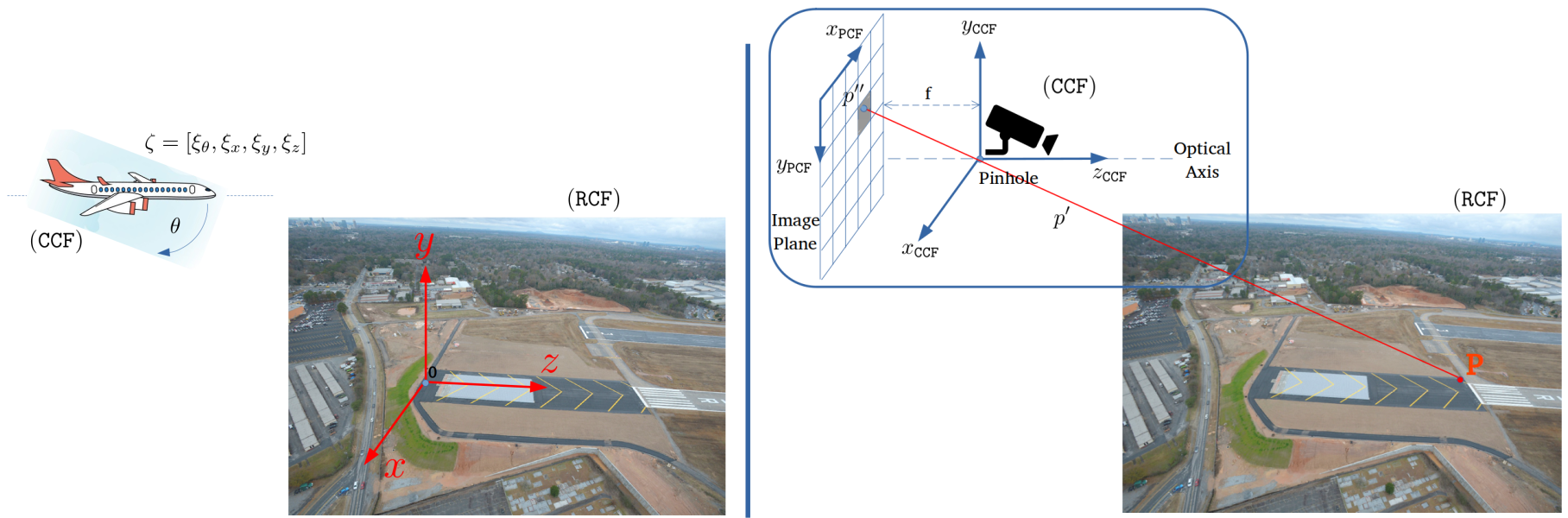}
\caption{Main coordinate frames: Runway ($\texttt{RCF}$), Camera ($\texttt{CCF}$) and Pixel ($\texttt{PCF}$).}
\label{coordinates}
\end{figure}

~\\
\noindent \textbf{Aircraft Dynamical Model.}
In this paper, we will consider an aircraft landing on a runway. We assume the states of the aircraft to be measured with respect to the origin of the Runway Coordinate Frame (shown in Figure~\ref{coordinates}(left)), where positions are: $\xi_x$ is the axis across runway; $\xi_y$ is the altitude, and $\xi_z$ is the axis along runway. We consider only one angle $\xi_\theta$ which represents the pitch rotation around $x$ axis of the aircraft. The state vector of the aircraft at time $t \in \mathbb{N}$ is denoted by $\xi^{(t)} \in \mathbb{R}^4 = [\xi_\theta^{(t)}, \xi_x^{(t)}, \xi_y^{(t)}, \xi_z^{(t)}]^T$ and is assumed to evolve over time while being governed by a general nonlinear dynamical system of the form $ \xi^{(t+1)}=f(\xi^{(t)}, u^{(t)})$
where $u^{(t)} \in \mathbb{R}^m$ is the control vector at time $t$. Such nonlinear dynamical system is assumed to be time-sampled from an underlying continuous-time system with a sample time equal to $\tau$. 


~\\
\noindent \textbf{Runway Parameters.} We consider runway that consists of two line segments $L$ and $R$. Each line segment can be characterized by its start and end point (measured also in the Runway Coordinate Frame) i.e. $L = [(L_x, 0, L_z), (L_x, 0, L_z + r_{l})]$ and $R = [(R_x, 0, R_z), (R_x, 0, R_z + r_l)]$, with $R_x = L_x + r_w$ and $R_z = L_z$ where $r_w$ and $r_l$ refers to the runway width and length (standard international runways are designed with $r_w = 40$ meters wide and $r_l= 3000$ meters).

~\\
\noindent \textbf{Camera Model.} We assume the aircraft is equipped with a monochrome camera $\mathcal{C}$ that produces an image $I$ of $q \times q$ pixels. Since the camera is assumed to be monochromatic, each pixel in the image $I$ takes a value of 0 or 1. The image produced by the camera depends on the relative location of the aircraft with respect to the runway. In other words, we can model the camera $\mathcal{C}$ as a function that maps aircraft states into images, i.e., $\mathcal{C}: \mathbb{R}^4 \rightarrow \mathbb{B}^{q\times q}$. Although the images created by the camera depend on the runway parameters, for ease of notation, we drop this dependence from our notation in $\mathcal{C}$.

We utilize an ideal pinhole camera model~\cite{ma2012invitation} to capture the image formation process of this camera. In general, a point $p = (p_x,p_y,p_z)$ in the Runway Coordinate Frame ($\texttt{RCF}$) is mapped into a point $p' = (p'_{x_\texttt{CCF}}, p'_{y_\texttt{CCF}}, p'_{z_\texttt{CCF}})$ on the Camera Coordinate Frame ($\texttt{CCF}$) using a translation and rotation transformations defined by~\cite{MultipleView}:
\begin{equation}
    \begin{bmatrix}
    p'_{x_\texttt{CCF}} \\
    p'_{y_\texttt{CCF}} \\
    p'_{z_\texttt{CCF}} \\
    1
\end{bmatrix}=\begin{bmatrix}
    1 & 0 & 0 & x\\
    0 & \cos{\theta} & \sin{\theta} & y\\
    0 & -\sin{\theta} & \cos{\theta} & z\\
    0 & 0 & 0 & 1
\end{bmatrix}\begin{bmatrix}
    p_x \\
    p_y \\
    p_z \\
    1
\end{bmatrix}
\end{equation}
The camera then converts the 3-dimensional point $p'$ on the camera coordinate frame into two-dimensional point $p''$ on the Pixel Coordinate Frame ($\texttt{PCF}$) as:
\begin{align}
    p'' = \left(p''_{x_\texttt{PCF}}, p''_{y_\texttt{PCF}} \right) = \left(\left\lfloor \frac{ q_{x_\texttt{PCF}} }{ q_{z_\texttt{PCF}}} \right\rfloor, \left\lfloor\frac{ q_{y_\texttt{PCF}} }{ q_{z_\texttt{PCF}}} \right\rfloor \right)
\end{align} 
where:
\begin{equation}
    \begin{bmatrix}
    q_{x_\texttt{PCF}} \\
    q_{y_\texttt{PCF}} \\
    q_{z_\texttt{PCF}}
\end{bmatrix}=
\begin{bmatrix}
\rho_w & 0 & u_0\\
0 & -\rho_h & v_0\\
0 & 0 & 1
\end{bmatrix}
\begin{bmatrix}
    f & 0 & 0 & 0\\
    0 &  f & 0 & 0\\
    0 & 0 & 1 & 0
\end{bmatrix}
\begin{bmatrix}
    p'_{x_\texttt{CCF}} \\
    p'_{y_\texttt{CCF}} \\
    p'_{z_\texttt{CCF}} \\
    1
\end{bmatrix}
\end{equation}
and $f$ is the focal length of the camera lens, $\textrm{W}$ is the image width (in meters), $\textrm{H}$ is the image width (in meters), $\textrm{WP}$ is the image width (in pixels), $\textrm{HP}$ is the image height (in pixels), and $u_0 = 0.5 \times \textrm{WP}$,$v_0 = 0.5 \times \textrm{HP}$, $\rho_w=\frac{\textrm{WP}} {\textrm{W}}$, $\rho_h=\frac{\textrm{HP}}{\textrm{H}}$. 

What is remaining is to map the coordinates of $p'' = \left(p''_{x_\texttt{PCF}}, p''_{y_\texttt{PCF}} \right)$ into a binary assignment for the different $q\times q$ pixels. But first, we need to check if $p''$ is actually inside the physical limits of the Pixel Coordinate Frame ($\texttt{PCF}$) by checking:
%
%
\begin{equation}
  \text{visible} =
    \begin{cases}
      \text{yes} & |p''_{x_\texttt{PCF}}| \leq \frac{W}{2} \ \vee \ |p''_{y_\texttt{PCF}}| \leq \frac{H}{2}\\
      \text{no} & \text{otherwise}
    \end{cases}       
\end{equation}
Whenever the point $p''$ is within the limits of $\texttt{PCF}$, then the pixel $I[i,j]$ should be assigned to 1 whenever the index of the pixel matches the coordinates $\left(p''_{x_\texttt{PCF}}, p''_{y_\texttt{PCF}} \right)$, i.e.:
\begin{equation}
  I[i,j] =
    \begin{cases}
      1 & ( p''_{x_\texttt{PCF}} == i-1) \wedge (p''_{y_\texttt{PCF}} == j-1) \wedge \text{visible}\\
      0 & \text{otherwise}
    \end{cases}  
\end{equation}
for $i,j \in (1,2,3...\textrm{WP})$. Where for simplicity, we set $\textrm{HP}=\textrm{WP}$ for square images. This process of mapping a point $p$ in the Runway Coordinate Frame $(\mathtt{RCF})$ to a pixel in the image $I$ is summarized in Figure~\ref{coordinates} (right).

~\\
\noindent \textbf{Neural Network Controller.} The aircraft is controlled by a vision based neural network $\mathcal{NN}$ controller that maps the images $I$ created by the camera $C$ into a control action, i.e., $\mathcal{NN}:  \mathbb{B}^{q\times q}  \rightarrow \mathbb{R}^m$. We confine our attention to neural networks that consist of multiple layers and where Rectified Linear Unit (ReLU) are used as the non-linear activation units.

~\\
\noindent \textbf{Problem Formulation.} Consider the closed-loop vision based system $\Sigma$ defined as:
$$ \Sigma: \begin{cases} \; \xi^{(t+1)} = f(\xi^{(t)}, \mathcal{NN}(\mathcal{C}(\xi^{(t)}))). \end{cases} $$
A trajectory of the closed loop system $\Sigma$ that starts from the initial condition $\xi_0$ is the sequence $\{\xi^{(t)}\}_{t = 0, \xi^{(0)} = \xi_0}^{\infty}$.
Consider also a set of initial conditions $\mathcal{X}_0 \subset \mathbb{R}^4$. We denote by $\Sigma^{\mathcal{X}_0}$ the trajectories of the system $\Sigma$ that starts from $\mathcal{X}_0$, i.e.,  
$$\Sigma^{\mathcal{X}_0} = \bigcup_{\xi_0 \in \mathcal{X}_0} \{\xi^{(t)}\}_{t = 0, \xi^{(0)} = \xi_0}^{\infty}.$$

We are interested in checking if the closed-loop system meets some specifications that are captured using Linear Temporal Logic (LTL) (or a Bounded-Time LTL) formula $\varphi$. Examples of such formulas may include, but are not limited to:
\begin{itemize}
    \item $\varphi_1 := \Diamond \{\xi_\theta = 0 \land \xi_y = 0\}$ which means that the aircraft should \emph{eventually} reach an altitude of zero while the pitch angle is also zero. Satisfying $\varphi_1$ ensures that the aircraft landed on the ground.
    \item $\varphi_2 := \Box \{ \xi_z \le 3000\}$ which ensures the aircraft will \emph{always} land before the end of the runway (assuming a runway length that is equal to 3000 meters).
\end{itemize}
For the formal definition of the syntax and semantics of  LTL and Bounded-Time LTL formulas, we refer the reader to~\cite{clarke2018handbook}. Given a formula $\varphi$ that specifies correct landing, our objective is to design a bounded model checking framework that verifies if all the trajectories $\Sigma^{\mathcal{X}_0}$ satisfy $\varphi$ (denoted by $\Sigma^{\mathcal{X}_0} \models \varphi$).

%% file: SEC4-Framework/framework.tex
\section{FRAMEWORK}

The verification problem described in Section 2 is challenging because it needs to take into account the nonlinear dynamics of the aircraft $f$, the image formation process captured by the camera model $\mathcal{C}$, and the neural network controller $\mathcal{NN}$.



\begin{figure}[tbh!]
\centering
\includegraphics[width=1.0\columnwidth]{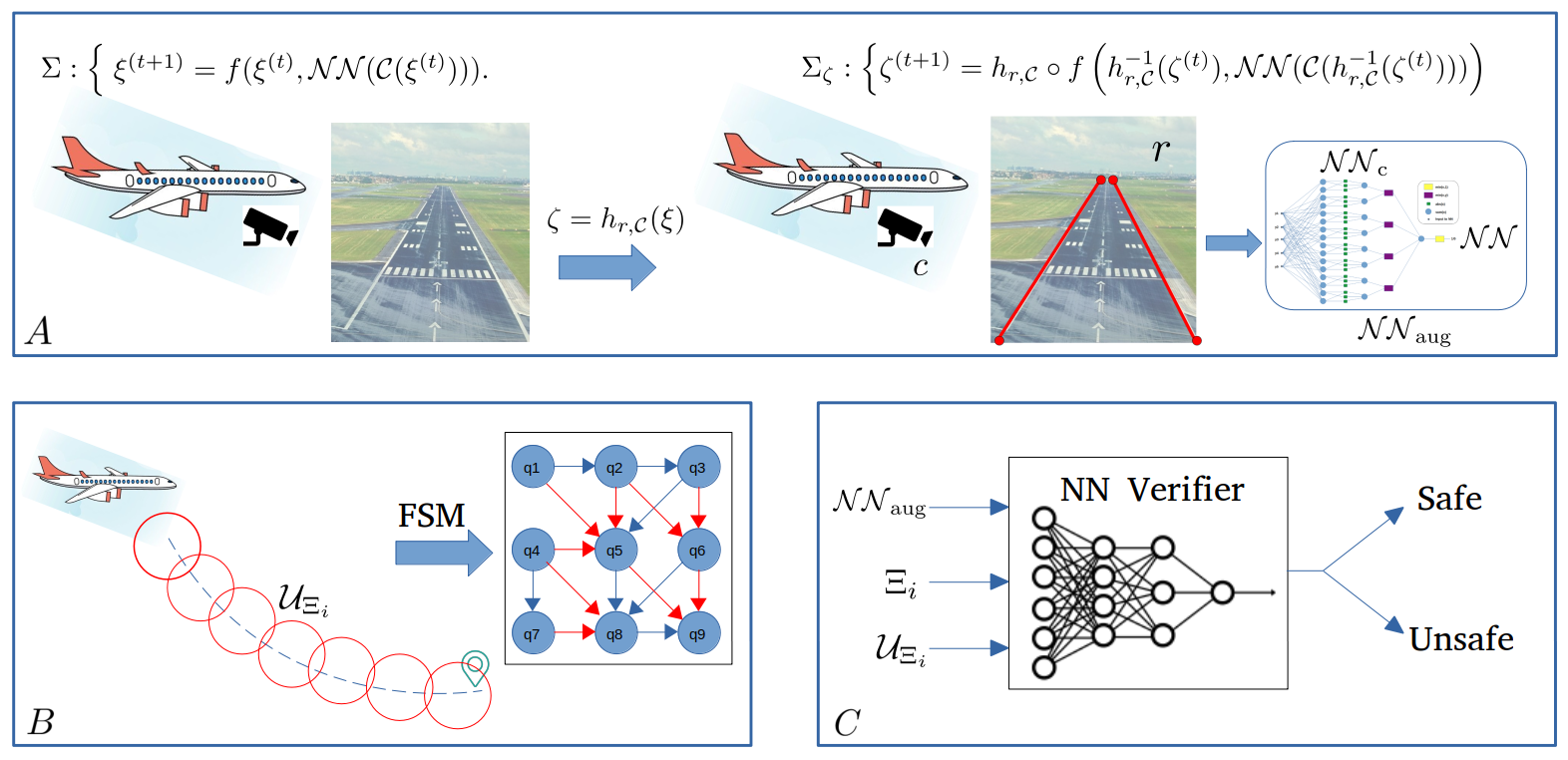}
\caption{Main elements of the proposed NNLander-VeriF framework: (A): construction of the augmented neural network that captures both perception and control, (B:) symbolic analysis of aircraft trajectories, (C:) neural network verification.}
\label{alrc_architecture_general}
\end{figure}

Our framework starts by re-modeling the pinhole camera model as a ReLU based neural network (with manually designed weights) that we refer to as the perception neural network $\mathcal{NN}_\mathcal{C}$. To facilitate this re-modeling, we need first to apply a change of coordinates to the states of the dynamical systems. We refer to the states in the new coordinates as $\zeta$, i.e., $\zeta = h(\xi)$. By augmenting $\mathcal{NN}_\mathcal{C}$ along with the neural network controller $\mathcal{NN}$, one can obtain an augmented neural network $\mathcal{NN}_{\text{aug}}:\mathbb{R}^n \rightarrow \mathbb{R}^m$ defined as $\mathcal{NN}_{\text{aug}} = \mathcal{NN} \circ \mathcal{NN}_C$ and a simplified closed-loop dynamics, in the new coordinates, written as:
$$ \Sigma: \begin{cases} \; \zeta^{(t+1)} = g(\zeta^{(t)}, \mathcal{NN}_{\text{aug}}(\zeta^{(t)})). \end{cases} $$

Now, assume that we are given (i) a region $\Xi$ in the new coordinate system and (ii) the maximal set of control actions (denoted by $\mathcal{U}_{\Xi}$) that can be applied at $\Xi$ while ensuring the system adhere to the specification $\varphi$. Given this pair $(\Xi, \mathcal{U}_{\Xi})$ one can always ensure that the augmented neural network $\mathcal{NN}_{\text{aug}}$ will produce actions in the set $\mathcal{U}_{\Xi}$ whenever its inputs are restricted to $\Xi$ by checking the following property:
\begin{align}
    \forall \zeta \in \Xi . \big(\mathcal{NN}_{\text{aug}}(\zeta) \in \mathcal{U}_{\Xi} \big)
    \label{eq:nn_prop}
\end{align} 
which can be easily verified using existing neural network model checkers~\cite{katz2019marabou,KhedrPEREGRiNNPenalizedRelaxationGreedy2020,ferlez2021fast}. In other words, checking the augmented neural network against the property above ensures that all the images produced within the region $\Xi$ will force the neural network controller $\mathcal{NN}$ to produce control actions that are within the set of allowable actions $\mathcal{U}_{\Xi}$.

To complete our framework, we need to partition the state-space into regions $(\Xi_1, \Xi_2, \ldots)$. Each region is a ball parametrized by a center $\zeta_i$ and a radius $\epsilon$. For each region, our framework will compute the set of allowable control actions at each of these regions $(\mathcal{U}_{\Xi_1}, \mathcal{U}_{\Xi_2}, \ldots)$. Our framework will also parametrize each set $\mathcal{U}_{\Xi_i}$ as a ball with center $c_i$ and radius $\mu_i$, i.e., $\mathcal{U}_{\Xi_i} = \{u \in \mathbb{R}^{4} | \; \Vert u - c_i \Vert \le \mu_i \}$. 
The computations of the pairs $(\Xi_i, \mathcal{U}_{\Xi_i})$ can be carried out using the knowledge of the aircraft dynamics $f$.
%
%
%
In summary, and as shown in Figure~\ref{alrc_architecture_general}, our framework will consist of the following steps:
\begin{itemize}
    \item \textbf{(A) Compute the augmented neural network}: Using the physical model of the pinhole camera, our framework will re-model the pinhole camera $\mathcal{C}$ as a neural network that can be augmented with the neural network controller to produce a simpler model that is amenable for verification.
    
    \item \textbf{(B) Compute the set of allowable control actions}:
    We use the properties of the dynamical system $f$ to compute the set of safe control actions $\mathcal{U}_{\Xi_i}$ for each partition $\Xi_i$ of the state space.
    \item \textbf{(C) Apply the neural network model checker:} We use the neural network model checkers to verify that $\mathcal{NN}_{\text{aug}}$ satisfies~\eqref{eq:nn_prop} for each identified pair $(\Xi, \mathcal{U}_{\Xi_i})$.
\end{itemize}

The remainder of this paper is devoted to providing details for the steps required for each of the three phases above.

\section{Neural Network Augmentation}
\begin{figure}[tbh!]
\centering
\includegraphics[width=1.0\columnwidth]{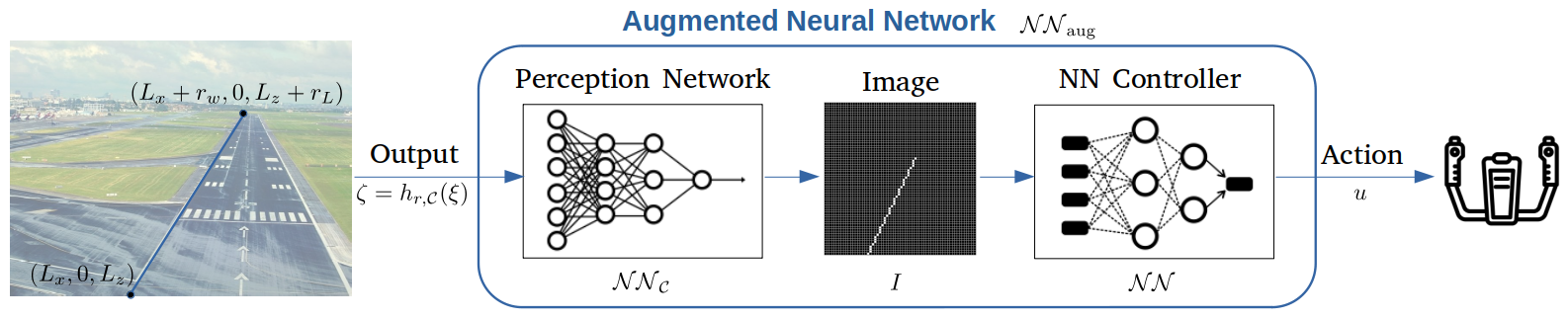}
\caption{Augmented network $\mathcal{NN}_{\text{aug}}$ maps the output $\zeta$ to control action $u$.}
\label{fig:augmented_nn}
\end{figure}

In this section, we focus on the problem of using the geometry of the runway to develop a different mathematical model for the camera $\mathcal{C}$. As argued in the previous section and shown in Figure~\ref{fig:augmented_nn}, our goal is to obtain a model with the same structure of a neural network (i.e., consists of several layers and neurons) and contains only ReLU activation units. We refer to this new model as $\mathcal{NN}_\mathcal{C}$.

The main challenge to construct $\mathcal{NN}_\mathcal{C}$ is the fact that ReLU based neural networks can only represent piece-wise affine (or linear) functions~\cite{nagamine2017understanding}. Nevertheless, the camera model $\mathcal{C}$ is inherently nonlinear due to the optical projection present in any camera. Such non-linearity can not be expressed (without any error) via a piece-wise affine function. To solve this problem, we propose a change of coordinates to the aircraft states $h$. Such change of coordinates is designed to eliminate part of the camera's non-linearity while allowing the remainder of the model to be expressed as a piece-wise affine transformation.



\noindent\textbf{Change of Coordinates:}
Recall the runway consists of line segments $L$ and $R$ (defined in Section 2). Instead of measuring the state of the aircraft by the vector $\zeta = [\xi_\theta,\xi_x,\xi_y,\xi_z]$, we propose measuring the state of the aircraft by the projections of the end points of the lines $L$ and $R$ on the Pixel Coordinate Frame $\texttt{PCF}$. Formally, we define the change of coordinates as:
\begin{align}
    \zeta &= h_{r,\mathcal{C}}  (\xi) =
    \begin{bmatrix} 
    \zeta_1 \\~\\ \zeta_2 \\~\\ \zeta_3 \\~\\ \zeta_4 \\~\\ \zeta_5
    \end{bmatrix}
    =
     \begin{bmatrix} 
    \rho_wf\frac{L_x+\xi_x}{L_z cos(\xi_\theta)+\xi_z}+u_0 
    \\~\\
      -\rho_hf\frac{L_z sin(\theta)+\xi_y}{L_z cos(\xi_\theta)+\xi_z}+v_0 
    \\~\\
       \rho_wf\frac{L_x +\xi_x}{(L_z + r_L) cos(\xi_\theta)+\xi_z}+u_0
    \\~\\
       -\rho_hf\frac{(L_z + r_L) sin(\xi_\theta)+\xi_y}{ (L_z + r_L) cos(\xi_\theta)+\xi_z}+v_0
    \\~\\
       \zeta_1 \zeta_4 - \zeta_2 \zeta_3
    \end{bmatrix}
\end{align}
where $f,\rho_h, \rho_w, v_0, u_0$ are the camera physical parameters as defined in Section 2. In other words, the pair $(\zeta_1, \zeta_2)$ is the projection of the start point of the runway $(L_x,0,L_z)$ onto the Pixel Coordinate Frame $\texttt{PCF}$ (while ignoring the flooring operator for now). Similarly, the pair $(\zeta_3, \zeta_4)$ is the projection of the endpoint of the runway $(L_x,0,L_z+r_L)$ onto the $\texttt{PCF}$ frame. Indeed, we can define a similar set of variables for the other line segment of the runway, $R$. The dependence of this change of coordinates on the camera parameters (e.g., the focal length $f$) and the runway parameters justifies the subscripts in our notation $h_{r,\mathcal{C}}$. We refer to the new state-space as $\Xi$.










Before we proceed, it is crucial to establish the following result.
\begin{proposition}
The change of coordinates function $h_{r,\mathcal{C}}$ is bijective.
\end{proposition}
The proof of such proposition is based on ensuring that the inverse function $h^{-1}_{r,\mathcal{C}}$ exists. For brevity, we will omit the details of this proof. Since $h_{r,\mathcal{C}}$ is bijective, we can re-write the closed-loop dynamics of the system as:
\begin{align}
    \Sigma_{\zeta}: \begin{cases} \zeta^{(t+1)} = h_{r,\mathcal{C}} \circ f\left( h^{-1}_{r,\mathcal{C}}(\zeta^{(t)}), \mathcal{NN}( \mathcal{C}(h^{-1}_{r,\mathcal{C}}(\zeta^{(t)}))) \right) \end{cases}
    \label{eq:sys_dyn_2}
\end{align} 
Indeed, if $\Sigma_{\zeta}$ satisfies the property $\varphi$ then do the original system $\Sigma$ and vice versa, thanks for the fact that $h_{r,\mathcal{C}}$ is bijective. This is captured by the following proposition:
\begin{proposition}
Consider the dynamical systems $\Sigma$ and $\Sigma_\zeta$. Consider a set of initial states $\mathcal{X}_0$ and an LTL formula $\varphi$, the following holds:
$$ \Sigma^{\mathcal{X}_0} \models \varphi \Longleftrightarrow \Sigma^{\Xi_0}_\zeta \models \varphi$$
where $\Xi_0 = \{h_{r,\mathcal{C}}(\xi) \; | \; \xi \in \mathcal{X}_0\}$.
\end{proposition}

~\\
\noindent\textbf{Neural Network-based Model for Perception:} 
While the model of the pinhole camera (defined in equation (1)-(5)) focuses on mapping individual points into pixels, we aim here to obtain a model that maps the entire runway lines $R$ and $L$ into the corresponding binary assignment for each pixel in the image. Therefore, it is insufficient to analyze the values of $\zeta_1,\ldots,\zeta_4$ which encodes the start point $(\zeta_1,\zeta_2)$ and the endpoint $(\zeta_3,\zeta_4)$ of the runway line segments on the \texttt{PCF}. To correctly generate the final image $I \in \mathbb{B}^{q \times q}$, we need to map \emph{every} point between $(\zeta_1,\zeta_2)$ and $(\zeta_3,\zeta_4)$ into the corresponding pixels. 

\begin{figure}[tbh!]
\centering
\includegraphics[width=0.8\columnwidth]{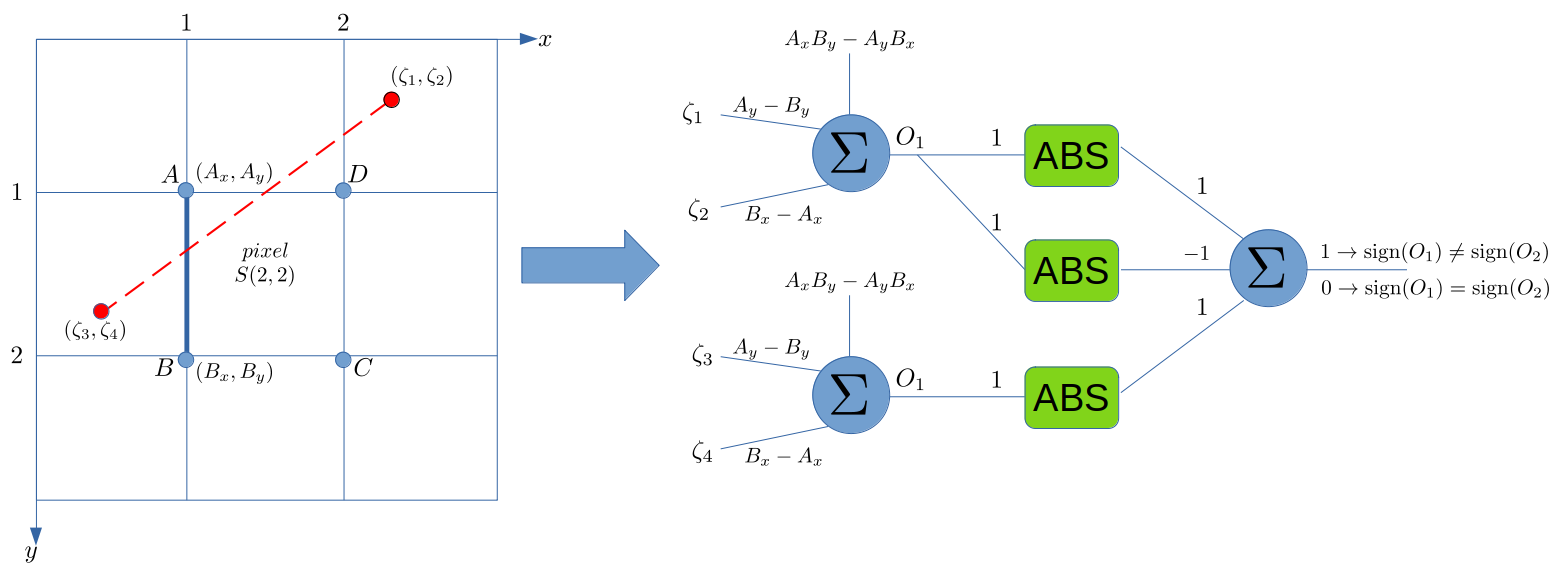}
\caption{Line-segment intersection algorithm: The runway line (in red) as seen by the camera intersects the pixel edge $A-B$ (in blue), this single edge intersection is detected by using a layer of six ReLU's.}
\label{fig:frompixtoneuron}
\end{figure}

While the pinhole camera (defined in equation (1)-(4)) uses the information in the Pixel Coordinate Frame $(\texttt{PCF})$ to compute the values of each pixel, we instead rely on the information in the Camera Coordinate Frame $(\texttt{CCF})$ to avoid the nonlinearities added by the flooring operator in (2) and the logical checks in (4)-(5). For each pixel, imagine a set of four line segments $AB, BC, CD, DA$ in the Pixel Coordinate Frame ($\texttt{PCF}$) that defines the edges of each pixel (see Figure~\ref{fig:frompixtoneuron} for an illustration). To check if a pixel should be set to zero or one, it is enough to check the intersection between the line segment $(\zeta_1,\zeta_2)-(\zeta_3,\zeta_4)$ and each of the lines $A-B, B-C, C-D, D-A$. Whenever an intersection occurs, the pixel should be assigned to one.




To intersect one of the pixel edges, e.g., the edge $A-B = (A_x, A_y)-(B_x,B_y)$, with the line segment $(\zeta_1,\zeta_2)-(\zeta_3,\zeta_4)$, we proceed with the standard line segment intersection algorithm~\cite{intersection} which compute four values named $O_1, O_2, O_3, O_4$ as:
\begin{align}
    O_1&=\zeta_1(A_y-B_y)+\zeta_2(B_x-A_x)+A_x B_y - A_y B_x \\
    O_2&=\zeta_3(A_y-B_y)+\zeta_4(B_x-A_x)+A_xB_y-A_yB_x\\
    O_3&=-\zeta_1(A_y)+\zeta_2(A_x)+\zeta_3(A_y)-\zeta_4(A_x)+\zeta_5\\
    O_4&=-\zeta_1(B_y)+\zeta_2(B_x)+\zeta_3(B_y)-\zeta_4(B_x)+\zeta_5
\end{align}
The line segment algorithm~\cite{intersection} detects an intersection whenever the following condition holds:
%
%
\begin{equation}
   (\text{sign}(O_1) \neq \text{sign}(O_2)) \ \wedge \ (\text{sign}(O_3) \neq sign(O_4)) 
\end{equation}

Luckily, we can organize the equations (9)-(13) in the form of a neural network with a Rectifier Linear Activation Unit (ReLU). ReLU nonlinearity takes the form of $\text{ReLU}(x) = \max \{x,0\}$. To show this conversion, we first note that the values of $A_x,A_y,B_x,B_y$ are constant and well defined for each pixel. So assuming the input to such a neural network is the vector $\zeta$, one can use equations (9)-(12) to assign the weights to the input layer of the neural network (as shown in Figure~\ref{fig:frompixtoneuron}). To check the signs of $O_1, \ldots, O_4$, we recall the well-known identity for numbers of the same sign:
\begin{equation}
    \text{sign}(a) = \text{sign}(b) \Longleftrightarrow |a+b| - |a|-|b| = 0
\end{equation}
The absolute function can be implemented directly with a ReLU using the identity:
\begin{equation}
    |x| = \max\{x,0\} + \max\{-x,0\}.
\end{equation}


\begin{figure}[tbh!]
\centering
\includegraphics[width=1.0\columnwidth]{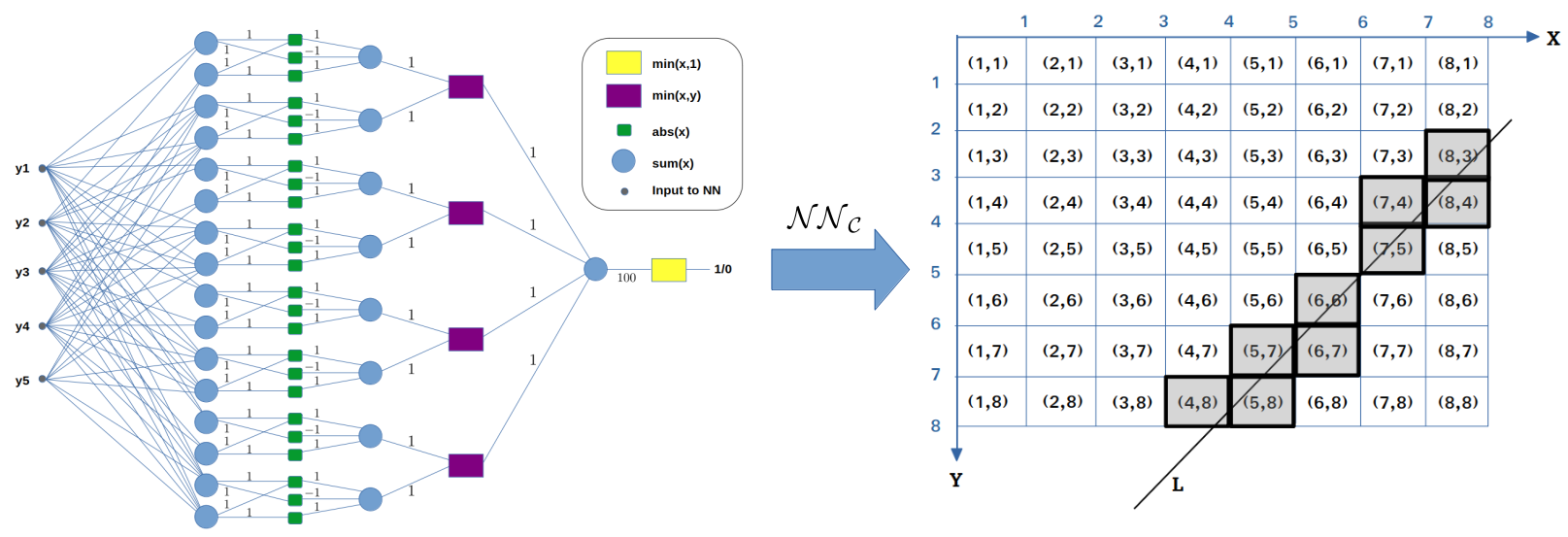}
\caption{$\mathcal{NN}_{\mathcal{C}}$ checks the intersection between line segment $(\zeta_1,\zeta_2)-(\zeta_3,\zeta_4)$ and all edges of each cell pixel of the final image.}
\label{fig:atest_cell}
\end{figure}

The process above has to be repeated four times (to account for all edges $A-B, B-C, C-D, D-A$ of a pixel). Finally, to check that at least one intersection occurred, we compute the minimum across the results from all the intersections. Calculating the minimum itself can be implemented directly with a ReLU using the identity:
\begin{equation}
    min\{a,b\} = \frac{a+b}{2} -  \frac{|a-b|}{2}.
\end{equation}
The overall neural network requires $68 \times q\times q$ ReLU neurons for each projected line segment. The final architecture is shown in Figure~\ref{fig:atest_cell}. We refer to the resulting neural network as $\mathcal{NN}_{\mathcal{C}}(\zeta^{(t)})$.

It is direct to show that the constructed neural network $\mathcal{NN}_{\mathcal{C}}(\zeta^{(t)})$ will produce the same images obtained by the pinhole camera model $\mathcal{C}$, i.e.,
$$\mathcal{C}(h^{-1}_{r,\mathcal{C}}(\zeta^{(t)})) = \mathcal{NN}_{\mathcal{C}}(\zeta^{(t)})$$
Finally, by substituting in~\eqref{eq:sys_dyn_2}, we can now re-write the closed-loop dynamics as:
\begin{align}
    \Sigma_{\zeta}: \begin{cases} 
    \zeta^{(t+1)} &= g\left( h^{-1}_{r,\mathcal{C}}(\zeta^{(t)}), \mathcal{NN}_{\text{aug}}(\zeta^{(t)}) \right)
    \end{cases}
    \label{eq:sys_dyn_3}
\end{align}
where $\mathcal{NN}_{\text{aug}} = \mathcal{NN}\circ\mathcal{NN}_{\mathcal{C}}$ and $g = h_{r,\mathcal{C}} \circ f$.

\section{Identifying the allowable control actions using symbolic abstractions
}
As shown in Section 3, our framework aims to split the verification of the dynamical system~\eqref{eq:sys_dyn_3} into several NN model checking queries. Each query will verify the correctness of the closed-loop system within a region (or a symbol) $\Xi_i$ of the state space. To prepare for such queries, we need to compute a set of input/output pairs $(\Xi_i, \mathcal{U}_{\Xi_i})$ with the guarantee that all the control inputs inside each $\mathcal{U}_{\Xi_i}$ will produce trajectories that satisfy the specifications $\varphi$. In this section, we provide details of how to compute the pairs $(\Xi_i, \mathcal{U}_{\Xi_i})$.

~\\
\noindent\textbf{State Space Partitioning:}
Given a partitioning parameter $\epsilon$, we partition the new coordinate space of $\zeta$ into $L$ regions $\Xi_1, \Xi_2, \ldots, \Xi_L$ such that each $\Xi_i$ is an infinity-norm ball with radius $\epsilon$ and center $c_i$. For simplicity of notation, we keep the radius $\epsilon$ constant within all the regions $\Xi_i$. However, the framework is generic enough to account for multi-scale partitioning schemes similar to those reported in the literature of symbolic analysis of hybrid systems~\cite{hsu2018multi}.

~\\
\noindent\textbf{Obtain Symbolic Models:}
Given the regions $\Xi_1, \Xi_2, \ldots$, the next step is to construct a \emph{finite-state abstraction} for the closed loop system~\eqref{eq:sys_dyn_3}. Such finite state abstraction takes the form of a finite state machine $\Sigma_q = (S_q, \sigma_q)$ where $S_q$ is the set of finite states and $\sigma_q: S_q \rightarrow 2^{S_q}$ is the state transition map of the finite state machine, defined as:
\begin{align}
    S_q = \{1, 2, \ldots L\} \quad \text{and} \quad
    j &\in \sigma_q(i) \Longleftrightarrow  g\left( h^{-1}_{r,\mathcal{C}}(c_i), \mathcal{NN}_{\text{aug}}(c_i) \right) \in \Xi_j.
\end{align}
In other words, the finite state machine (FSM) has a number of states $L$ that is equal to the number of regions $\Xi_i$, i.e., each finite state symbolically represents a region. A transition between the state $i$ and $j$ is added to the state transition map $\sigma_q$  whenever applying the NN controller to the center of the region $i$ (i.e, $c_i$) will force the next state of the system to be within the region $\Xi_j$. The value of $g\left( h^{-1}_{r,\mathcal{C}}(c_i), \mathcal{NN}_{\text{aug}}(c_i) \right)$ can be directly computed by evaluating the neural network $\mathcal{NN}_{\text{aug}}$ at the center $c_i$ followed by evaluating the function $g$.

So far, the state transition map $\sigma_q$ accounts only for actions taken at the center of the region. To account for the control actions in all the states $\zeta_i \in \Xi_i$, we need to bound the distance between the trajectories that start at the center of the region $c_i$ and the trajectories that start from any other state $\zeta_i \in \Xi_i$. For such bound to exist, we enforce an additional assumption on the dynamics of the aircraft model $f$ (and hence $g = h_{r,\mathcal{C}} \circ f$) named $\delta$ forward complete ($\delta$-FC)~\cite{fsm}. Given the center of a region $c_i$ and an arbitrary state $\zeta_i \in \Xi_i$, the $\delta$-FC assumption bounds the distance, denoted by $\delta_\zeta$, between the trajectories that starts at $\zeta_i$ and the center $c_i$ as:
\begin{equation}
    \delta_\zeta \leq \beta(\epsilon,\tau) + \gamma(||\mathcal{NN}_{\text{aug}}(c_i)-\mathcal{NN}_{\text{aug}}(\zeta_i)||_{\infty},\tau)
    \label{eq:delta_fc_bound}
\end{equation}
where $\tau$ is the sample time used to obtain the dynamics $f$ (as explained in Section 2) and $\beta$ and $\gamma$ are class $K_{\infty}$ functions that can be computed from the knowledge of the dynamics $f$. Such $\delta$-FC assumption is shown to be mild and does not require the aircraft dynamics to be stable. For technical details about the $\delta$-FC assumption and the computation of the functions $\beta$ and $\gamma$, we refer the reader to~\cite{PhdZamani}. Given the inequality~\eqref{eq:delta_fc_bound}, we can revisit the definition of the state transition map $\sigma_q$ to account for all possible trajectories as:
\begin{align}
    j &\in \sigma_q(i) \Longleftrightarrow  g\left( h^{-1}_{r,\mathcal{C}}(c_i), \mathcal{NN}_{\text{aug}}(c_i) \right) + \delta_\zeta \in \Xi_j.
\end{align}
With such a modification, it is direct to show the following result:
\begin{proposition}
Consider the dynamical systems $\Sigma_\zeta$ and $\Sigma_q$. Consider also a set of initial conditions $\Xi_0$ and a specification $\varphi$. The following holds:
$$ \Sigma_q^{\mathcal{S}_0} \models \varphi \Rightarrow \Sigma_\zeta^{\Xi_o} \models \varphi $$
where $\mathcal{S}_0 = \{ i \in \{1,\ldots,L \} \; | \; \exists \;  \zeta_0 \in \Xi_0: \;\;  \zeta_0 \in \Xi_i 
\}$.
\end{proposition}
This proposition follows directly from Theorem 4.1 in \cite{fsm}.


%

~\\
\noindent\textbf{Compute the set of allowable control actions:}
Unfortunately, computing the norm $||\mathcal{NN}_{\text{aug}}(c_i)-\mathcal{NN}_{\text{aug}}(\zeta_i)||_{\infty}$ (and hence $\delta_\zeta$) is challenging. As shown in~\cite{kallus2019assessing}, computing such norm is NP-hard and existing tools in the literature focus on computing an upper bound for such norm. Nevertheless, the bounds given by the existing literature constitute large error margins that will render our approach severely conservative. 

To alleviate the problem above, we use the inequality~\eqref{eq:delta_fc_bound} in a ``backward design approach''. We first search for the maximum value of $\delta_\zeta$ that renders $\Sigma_q$ compatible with the specification. To that end, we substitute the norm  $||\mathcal{NN}_{\text{aug}}(c_i)-\mathcal{NN}_{\text{aug}}(\zeta_i)||_{\infty}$  with a dummy variable $\mu$. By iteratively increasing the value of $\mu$, we will obtain different $\Sigma_q$, one for each value of $\mu$. We use a bounded model checker for each value of $\mu$ to verify if the resulting $\Sigma_q$ satisfies the specification. We keep increasing the value of $\mu$ until the resulting $\Sigma_q$ no longer satisfies $\varphi$. We refer to this value as $\mu_{\text{max}}$. What is remaining is to ensure that the neural network indeed respects the bound: 
$$ ||\mathcal{NN}_{\text{aug}}(c_i)-\mathcal{NN}_{\text{aug}}(\zeta_i)||_{\infty} \le \mu_{\text{max}} $$
To that end, we define the set of allowable control actions $\mathcal{U}_{\Xi_i}$ as:
$$\mathcal{U}_{\Xi_i} = \mathcal{B}_{\mu_{\text{max}}} (\mathcal{NN}_{\text{aug}}(c_i))$$
It is then direct to show the following equivalence:
$$ ||\mathcal{NN}_{\text{aug}}(c_i)-\mathcal{NN}_{\text{aug}}(\zeta_i)||_{\infty} \le \mu_{\text{max}} \Longleftrightarrow \forall \zeta \in \Xi_i . \big(\mathcal{NN}_{\text{aug}}(\zeta) \in \mathcal{U}_{\Xi_i} \big) $$
where $\mathcal{U}_{\Xi_i} = \mathcal{B}_{\mu_{\text{max}}} (\mathcal{NN}_{\text{aug}}(c_i))$.
Luckily, the right-hand side of this equivalence is precisely what neural network model checkers are capable of verifying. Algorithm~1 summarizes this discussion. The following result captures the guarantees provided by the proposed framework:
\begin{proposition}
The LanderNN-VeriF algorithm (Algorithm 1) is sound but not complete.
\end{proposition}

\begin{algorithm}[t!]
\caption{LanderNN-VeriF}  \label{alg:PolyARBerNN}
\begin{flushleft}
\textbf{Input:} $\Xi, \Xi_0, \varphi$, $\epsilon$, $\tau$, $\beta$, $\gamma$,  $\mathcal{NN}_{\text{aug}}$, $T$, $\overline{\mu}$, $\underline{\mu}$, $f$, $h$, $h^{-1}$\\
\textbf{Output}: $\texttt{STATUS}$
\end{flushleft}
\begin{algorithmic}[1]
\STATE $\{\Xi_1, \Xi_2, \ldots, \Xi_L\} = \texttt{Partition\_into\_regions}(\Xi, \epsilon)$
\STATE $\mu=\underline{\mu}$
\WHILE{\texttt{statusFSM} == UNSAT}
    \STATE $\Sigma_q = \texttt{Create\_FSM}(f,h,h^{-1},\tau,\beta,\gamma,  \mathcal{NN}_{aug},\Xi_{1..L}, \mu)$\\
    \STATE $\texttt{statusFSM} = \texttt{Check\_FSM} (\varphi, \Sigma_q, T)$\\
    \IF{$\mu \leq \overline{\mu}$}
        \STATE $\mu=\texttt{Increase\_MU}(\mu)$\\
    \ENDIF
\ENDWHILE
\FOR{i = 1 to L}
 \STATE $\texttt{STATUS\_NN[i]} = \texttt{NN\_Verifier}(\mathcal{NN}_{\text{aug}},\Xi_{i}, \mu)$\\
 \IF{$\texttt{STATUS\_NN[i]}==\text{SAT}$}
    \STATE \texttt{STATUS} = UNSAFE\\
 \ELSE
    \STATE \texttt{STATUS} = SAFE\\
 \ENDIF
\ENDFOR
\RETURN \texttt{STATUS}
\end{algorithmic}
\end{algorithm}

%% file: SEC6-Numerical/example.tex
\section{NUMERICAL EXAMPLE}

We illustrate the results in this paper using a vision-based aircraft landing system. We use a fixed-wing aircraft model defined using the guidance kinematic model \cite{UAV}, where orientations (in Rads) are defined by the course angle $\chi$ (rotation around $y_\texttt{CCF}$ axis), pitch angle $\theta$ (rotation around $x_\texttt{CCF}$ axis) and $V_g$ denotes the total Aircraft velocity relative to the ground. We further simplify the system by keeping the course angle pointing towards the runway ($\chi=0$), similarly velocity is kept as constant. Moreover, $\dot \theta$ (Rad/s) is regarded as the control input $u$. According to this model, the state vector of the aircraft
evolves over time while being governed by the following dynamical system \cite{UAV}:
%
%
%
%
%
\begin{align}
    \xi_z^{(t+1)} &=\xi_z^{(t)}+V_g \tau \cos{(\xi_\theta^{(t)})} \label{eq:aircraft_dyn_1} \\
    \xi_y^{(t+1)} &=\xi_y^{(t)}+V_g \tau \sin{(\xi_\theta^{(t)})} \\
    \xi_\theta^{(t+1)} &=\xi_\theta^{(t)}+ u^{(t)} \tau \label{eq:aircraft_dyn_3}
\end{align}
where $\tau$ is the sampling time.
For our simulations we consider $V_g=25\frac{m}{s}$ and $\tau=0.1$. Moreover based on airport standards we consider the runway segments (in meters) defined by $L = [(L_x, 0, L_z), (L_x, 0, L_z + r_{l})]$ and $R = [(R_x, 0, R_z), (R_x, 0, R_z + r_l)]$ where $R_x=20$, $L_x=-20$, $R_z=0$, $L_z=0$, $r_{l}=3000$.   For the camera parameters we consider images of $16 \times 16$ pixels and focal length of $400 \ mm$.

We note that the system dynamics~\eqref{eq:aircraft_dyn_1}-\eqref{eq:aircraft_dyn_3} is a $\delta$-FC system. In particular, by using the method \cite{PhdZamani} and the $\delta$-FC Lyapunov function $\mathcal{V}(\xi,\xi')= ||\xi-\xi'||_2^2$ one can show that:

\begin{equation}
    \beta(\zeta_1,\zeta_2,\zeta_3,\tau) = \sqrt{8} \ \sqrt{\zeta_1^2+\zeta_2^2+\zeta_3^2} \ e^{\tau}
\end{equation}
\begin{equation}
    \gamma(\mu,\tau) = \sqrt{V_g(e^{2\tau}-1)}\mu
\end{equation}

We work on the output space set $D=[\zeta_1\times \zeta_2 \times \zeta_3]=[0,16]\times[0,16]\times[0,16]$ of $\Sigma_{\zeta}$ with a precision $\epsilon=1$, thus our discretized grid consists of $16^3$ cubes. 

We used Imitation Learning to train a fully connected ReLU Neural Network controller ($\mathcal{NN}$) of 2 layers with 128 Neurons each. Trajectories from different initial conditions were collected and used to train the network.
Our objective is to verify that the aircraft landing using the trained $\mathcal{NN}_{\text{aug}}$ satisfies the safety specification $\phi= \square \lnot q_{\text{unsafe}}$ where $q_{\text{unsafe}} = [\xi_z=800, \xi_y=200, \xi_\theta=1]$ which corresponds to an unsafe region while landing.
\par
\begin{figure}[tbh!]
\centering
\includegraphics[width=1.0\columnwidth]{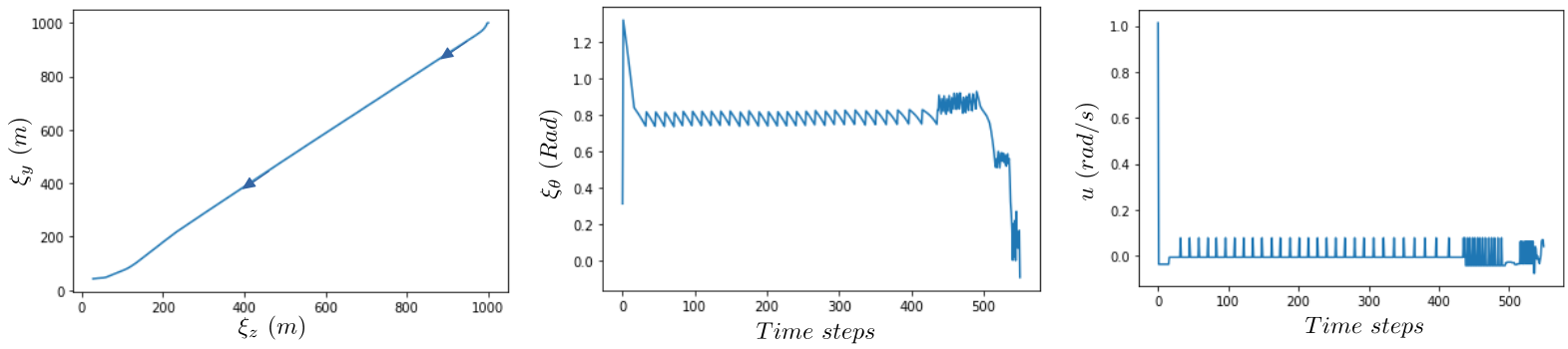}
\caption{Aircraft landing using augmented controller $\mathcal{NN}_{aug}$. Left: aircraft position ($\xi_y,\xi_z$); Middle: aircraft angle ($\xi_\theta$); Right: aircraft control ($u=\mathcal{NN}_{aug}$).}
\label{trajectory_controller}
\end{figure}

\begin{figure}[tbh!]
\centering
\includegraphics[width=1.0\columnwidth]{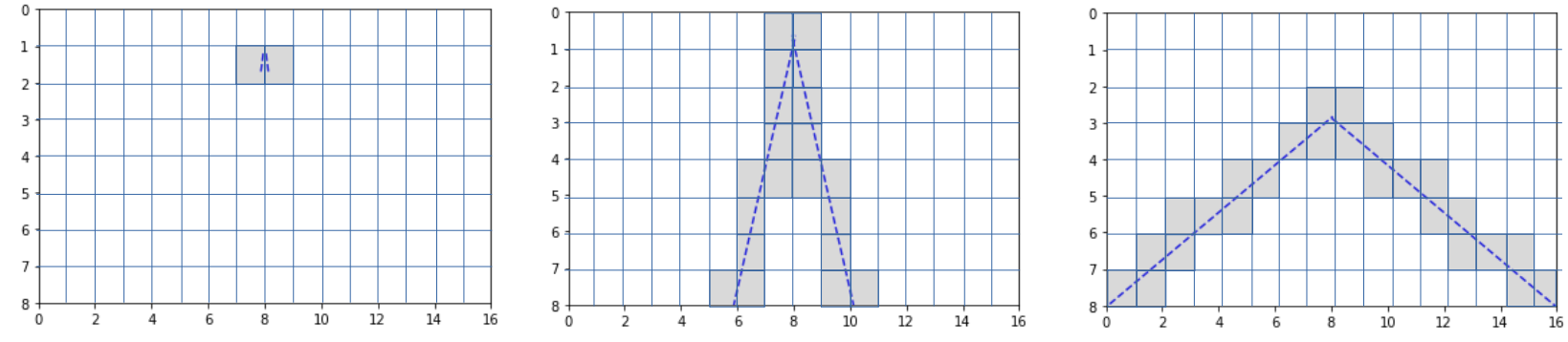}
\caption{Landing camera view using $16\times16$ pixels resolution. Left: $\xi^1=[1000,1000,\frac{\pi}{4}]$, Middle: $\xi^{300}=[400,300,\frac{\pi}{8}]$, Right: $\xi^{1000}=[5,5,0]$.}
\label{camera_view}
\end{figure}

 In what next, we report the execution time to verify the trained network. All experiments were executed on 
an Intel Core i7 processor with 50 GB of RAM. First, we implemented our Vision Network ($\mathcal{NN}_\mathcal{C}$) for images of $16\times16$ pixels using Keras.
Similarly, we used Keras composition libraries to merge the controller and perception networks into the augmented network ($\mathcal{NN}_{\text{aug}}$), a landing trajectory using $\mathcal{NN}_{aug}$ is shown in Figure~\ref{trajectory_controller} and its corresponding camera view is shown in Figure~\ref{camera_view}.

We used a Boolean SAT solver named SAT4J~\cite{sat4j} to implement the $\texttt{Check\_FSM}$ function in Algorithm 1. The finite state machine $\Sigma_q$ was encoded using a set of Boolean variables and our implementation performed a bounded model checking for the generated FSMs (the bounded model checking horizon was set to $20$). We constructed FSMs with the following values $\mu=[0.1,0.2,0.3,0.6,0.8,0.9,1.1]$ until a value of $\mu_{max}=1.1$ was found. The execution time for creating $\Sigma_q$ and verifying its properties with the bounded model checker increased monotonically from $2000$ seconds for $\mu = 0.1$ to $7000$ seconds for $\mu=1.1$.
As expected, the higher the value of $\mu$, the higher the number of transitions in $\Sigma_q$, and the higher the time needed to create and verify.

\begin{figure}[tbh!]
\begin{minipage}[c]{0.46\linewidth}
\includegraphics[width=\linewidth]{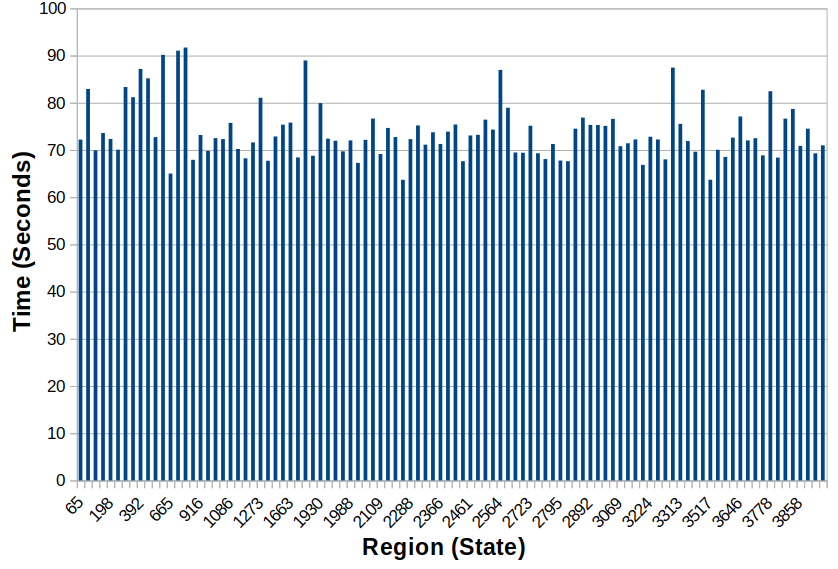}
\caption{Execution time for verifying $\varphi$ in 100 different random regions.}
\label{NNVerif}
\end{minipage}
\hfill
\begin{minipage}[c]{0.5\linewidth}
\includegraphics[width=\linewidth]{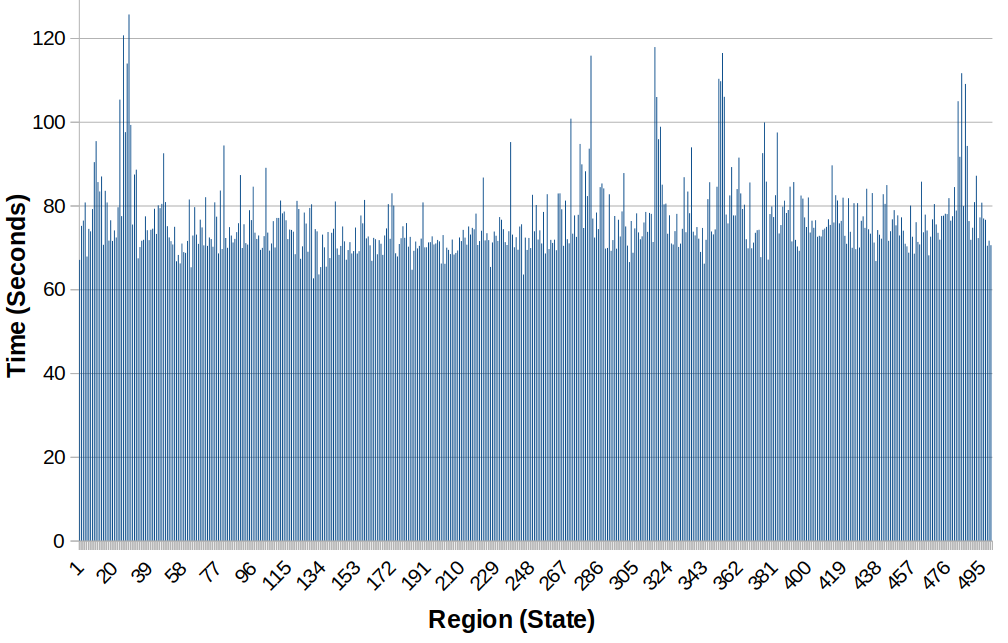}
\caption{Execution time for verifying $\varphi$ in regions 1 to 500.}
\label{NNVerif2}
\end{minipage}%
\end{figure}

Finally, we used PeregriNN~\cite{KhedrPEREGRiNNPenalizedRelaxationGreedy2020} as the NN model checker. Figure~\ref{NNVerif} reports the execution time for verifying the neural network property in 100 random regions, and Figure~\ref{NNVerif2} in regions 1 to 500. The average execution time was~76 seconds per region and the NN was found to be safe and satisfying the specification $\varphi$.






%% file: SEC7-Conclusion/conclusion.tex
\section{CONCLUSION AND FUTURE WORK}

Due to the recent surge in vision-based autonomous systems, it is becoming increasingly important to provide frameworks to facilitate its formal verification. 
In this work we have proposed two key contributions: first, a generative model that encodes part of the camera image formation process into a ReLU neural network, where the neuronal weights are fully determined by the camera intrinsic parameters, and second, a framework that uses the characteristics of the dynamical system (i.e. $\delta$-FC) to compute the set of safe control actions; Finally, having both contributions allows us to use off-the-shelf neural network checkers to verify the entire system.

At the same time, there are some limitations. 
First, the generative model we developed insists on modeling the image formation process with a piece-wise affine (PWA) function which facilitates encoding it as a ReLU network. However, this restriction may in odds with realistic scenarios which may not be captured exactly by CPWA functions. Nevertheless, it is widely known that CPWA functions can approximate general nonlinear functions with some error. 
This also leads to the second limitation, namely, the inability to consider noise in the image formation process.
Finally, the number of pixels has a direct effect on the scalability of the framework, as a consequence further improvements are required to build more concise finite-state machine abstractions of the physical system.

Moving forward, we plan to extend our approach in different directions to account for the aforementioned limitations. First, we seek to generalize the framework to account for uncertainties in the camera model, the image formation model, and the environment.
Second, we intend to process more complex image features (e.g. combinations of multiple lines and curvatures) by developing better generative models with provable error bounds. Finally, we aim to verify the robustness of neural network controllers to external disturbances (e.g., wind) while developing better scalable algorithms.
